\documentclass[]{template}

\usepackage{booktabs}
\usepackage{multirow}
\usepackage{colortbl}
\usepackage{multicol}
\usepackage{nicefrac}
\usepackage{latexsym}
\usepackage{float}
\usepackage{subcaption}
\usepackage{wrapfig}
\usepackage{bm}
\usepackage{tabularx}
\usepackage{ragged2e}
\newcolumntype{L}{>{\RaggedRight\hangafter=1\hangindent=0em}X}

\usepackage{mathtools}
\usepackage[linesnumbered,ruled,vlined]{algorithm2e}
\usepackage[capitalize,noabbrev]{cleveref}
\usepackage{calligra}
\DeclareMathAlphabet{\mathcalligra}{T1}{calligra}{m}{n}

\hypersetup{
    colorlinks=true,
    linkcolor=red,
    citecolor=Cerulean,
    filecolor=magenta,
    urlcolor=magenta,
}

\crefname{section}{§}{§§}
\Crefname{section}{§}{§§}

\theoremstyle{plain}

\theoremstyle{definition}

\theoremstyle{remark}

\renewcommand{\paragraph}[1]{\vspace{1mm}\noindent\textbf{#1}}
\DeclareCaptionLabelFormat{cont}{#1~#2\alph{ContinuedFloat}}
\usepackage{xspace}
\newcommand{\method}{\textsc{SERL}\xspace}

\newcommand\blfootnote[1]{%
  \begingroup
  \renewcommand\thefootnote{}\footnote{#1}%
  \addtocounter{footnote}{-1}%
  \endgroup
}

\setboolean{logo}{true}
\title{What and When to Distill: Selective Hindsight Distillation for Multi-Turn Agents}

\author[1,2]{Xiaozhe Li *}
\author[1,2]{Tianyi Lyu *}
\author[5]{Yang Li}
\author[2,3]{Yichuan Ma}
\author[2,3]{Peiji Li}
\author[2,3,4]{Linyang Li$^{\dag}$}
\author[2,3]{Qipeng Guo$^{\dag}$}
\author[2,4]{Dahua Lin}
\author[2,4]{Kai Chen}

\affil[1]{Tongji University}
\affil[2]{Shanghai AI Laboratory}
\affil[3]{Fudan University}
\affil[4]{The Chinese University of Hong Kong}
\affil[5]{Independent}

\begin{abstract}
Reinforcement learning can train LLM agents from sparse task rewards, but
credit assignment in long-horizon interactions remains a bottleneck: a
single success-or-failure signal must be distributed across dozens of
actions, most of which have no causal effect on the outcome. Existing
agent RL methods primarily rely on trajectory-level rewards or learned
proxy signals, without fully exploiting the per-step feedback that
environments naturally produce. Distillation-based approaches have been
combined with RL for single-turn reasoning, but the multi-turn agent
setting is underexplored---and fundamentally different. In an interactive
environment, feedback takes many forms: an error message after a failed
action, a changed page after a click, a new observation after navigation,
or even a successful reference trajectory. Which of these signals is most
useful for credit assignment, and where in a long trajectory each should
be applied, remain open questions. We conduct the first systematic study
of this design space for agent RL, spanning five feedback sources and two
insertion granularities. Guided by the findings, we introduce \method{},
a selective environment-reweighted learning framework whose key principle
is that \emph{the task reward determines the update direction, while
environment feedback adjusts only the placement and magnitude of that
update}. \method{} uses an environment-conditioned teacher to selectively
sharpen the RL objective on the actions that matter, without the
instability of unconstrained distillation. On ALFWorld and WebShop,
\method{} achieves $90.0\%$ and $80.1\%$ success respectively,
outperforming strong RL and distillation baselines. Our analysis reveals
that effective feedback is not simply the richest: grounded,
action-relevant signals at semantically meaningful insertion points
consistently outperform indiscriminate use of longer or more privileged
context.
\end{abstract}

\begin{document}

\blfootnote{$*$ Equal contribution.}
\blfootnote{$\dagger$ Corresponding authors: Linyang Li (lilinyang@pjlab.org.cn), Qipeng Guo (guoqipeng@pjlab.org.cn)}
\blfootnote{Code is at: \url{https://github.com/OliverLeeXZ/SERL}}

\maketitle

\section{Introduction}

Large language models are increasingly deployed as interactive agents that browse websites~\citep{webshop}, call tools~\citep{toolrl}, solve software engineering tasks~\citep{swebench}, perform machine learning workflows~\citep{mlebench,li2025opt}, and act in embodied environments~\citep{alfworld}.
Reinforcement learning is a natural training paradigm for these agents because many tasks provide a verifiable success signal at the end of an episode.
Yet the central difficulty is not obtaining the reward but \emph{assigning} it: a typical ALFWorld trajectory contains a dozen actions, of which only two or three change the environment in a meaningful way, while the rest are routine navigation or formatting steps.
When GRPO~\citep{grpo} or similar group-relative methods broadcast a single trajectory-level advantage to every token, high-leverage decisions and inert interface tokens receive the same update.
Better variance-reduction techniques can shrink gradient noise, but they cannot identify \emph{which} decision caused a later success or failure.

Agent environments already contain a dense signal that, in principle, addresses this problem.
After every action the environment returns feedback---an error message, an updated page, a changed object state---that reveals the local consequence of that specific decision.
This per-step feedback is far more informative than a single end-of-episode reward, which makes it a natural basis for credit assignment.
One way to exploit it is to let a teacher observe the environment's response to each action and use its token-level probabilities to supervise the student along the student's own rollout.
This idea, known as on-policy self-distillation (OPSD)~\citep{opd}, provides dense token-level supervision but introduces a new risk: the teacher inevitably conditions on \emph{privileged} information---post-action feedback, future observations, or successful reference trajectories---that the student cannot access at decision time.
Imitating the teacher indiscriminately therefore risks leaking unavailable information into the training target, amplifying stylistic preferences unrelated to task success, and destabilizing learning as the student policy drifts.

The question, then, is not \emph{whether} to use environment feedback but \emph{how}.
Two recent lines of work offer partial answers: SDPO~\citep{sdpo} adds a distillation loss to the policy objective, and RLSD~\citep{rlsd} uses teacher--student probability gaps to reweight the RL update.
However, both have been studied primarily in single-turn reasoning tasks where the feedback structure is simple and uniform.
Long-horizon agent training introduces two design axes that these methods do not address.
First, \textit{what} should the teacher see?
Environment feedback ranges from the immediate action response to the full future trajectory or a successful reference rollout; richer context gives the teacher more information but also more privilege.
Second, \textit{where} should the feedback affect learning?
It can be injected at every transition (step-level) or only at semantically meaningful state changes (anchor-level); the right granularity depends on how noisy and redundant the raw signal is.

We conduct the first systematic study of this design space for multi-turn agent RL, varying five feedback sources and two insertion granularities across ALFWorld and WebShop.
The study yields a clear finding: effective feedback is not simply the richest feedback---grounded, action-relevant signals at semantically meaningful insertion points consistently outperform indiscriminate use of longer or more privileged context.
Guided by this finding, we propose \method{}, a selective environment-reweighted learning framework built on an asymmetric principle: \textbf{the task reward determines the update direction; environment feedback only adjusts the placement and magnitude of that update.}
An environment-conditioned teacher scores the student's own action tokens with and without hindsight feedback; the resulting log-probability gap is converted into a bounded, sign-aware reweight of the GRPO advantage.
Distillation is restricted to executable action spans---reasoning and formatting tokens remain under the sole control of the reward-driven objective---and the teacher signal is decayed over training to prevent late-stage privileged-information leakage.

Our contributions are as follows:
\begin{itemize}[leftmargin=*,topsep=2pt,itemsep=1pt]
    \item We conduct the first systematic study of environment feedback for long-horizon LLM agent RL, analyzing how the \emph{source} and \emph{placement} of feedback jointly affect training stability and task performance across five feedback types and two insertion granularities.

    \item Guided by the study, we propose \method{}, a GRPO-compatible objective that converts privileged hindsight into a bounded, action-level reweight of the policy-gradient update, providing dense credit assignment while keeping the optimization direction anchored to the task reward.

    \item Experiments on ALFWorld and WebShop show that \method{} achieves $90.0\%$ and $80.1\%$ success respectively, outperforming strong RL and RL--distillation baselines. Our analysis reveals that grounded, action-relevant signals at semantically meaningful insertion points yield the strongest and most stable training.
\end{itemize}
\section{Related Work}
\textbf{Reinforcement learning for long-horizon LLM agents.}
Reinforcement learning has become a central post-training tool for LLMs, from RLHF~\citep{rlhf} to recent verifiable-reward training for reasoning and tool use~\citep{team2025kimi,guo2025deepseek,toolrl}. To reduce the cost of value modeling, critic-free and group-relative methods, such as RLOO~\citep{rloo}, GRPO~\citep{grpo}, DAPO~\citep{yu2025dapo}, and GSPO~\citep{gspo}, estimate advantages from multiple samples per query, enabling scalable training beyond PPO~\citep{schulman2017proximal}. These methods have shown strong performance in mathematical~\citep{guo2025deepseek}, logical~\citep{xie2025logic}, and optimization reasoning~\citep{npengine}.

With the increasing capability of LLMs and RL algorithms, LLM agents have shown strong potential in long-horizon, dynamic, and open-ended environments, including web navigation~\citep{webshop,appworld}, embodied tasks~\citep{alfworld}, search~\citep{searchr1}, and software engineering~\citep{swebench}. These long-horizon tasks introduce new challenges for RL, as success often depends on multi-turn interactions, delayed rewards, and environment-dependent decisions. Recent methods extend policy optimization to agentic settings by applying GRPO at the trajectory level over multi-turn rollouts~\citep{contextRL} or by performing stepwise policy optimization~\citep{wang2025ragen}. GIGPO~\citep{gigpo} and HGPO~\citep{hgpo} further exploit the hierarchical structure of agent trajectories, estimating advantages over actions, groups, or sub-trajectories to improve credit assignment. However, these RL methods still underutilize the rich feedback produced by agent environments, which can provide important signals for guiding LLM agent training.

\textbf{Credit assignment in long-horizon LLM agentic training.}
Credit assignment is a central challenge in agentic RL training. In methods such as GRPO~\citep{grpo}, the verifier usually provides only a sequence-level reward, so every token in a rollout receives the same advantage, regardless of whether it reflects a key decision or a stylistic filler. This is especially coarse for LLM agents, where final success depends on many intermediate states, actions, observations, and tool interactions.
Existing work improves credit granularity through process rewards, value models, and intermediate evaluators, which provide denser supervision for partial reasoning steps or action traces~\citep{lightman,luo2024improve,stepmath,zhang2024generative,cui2025process}. Other methods use token-level proxies such as entropy, uncertainty, attention, or outcome sensitivity to adjust updates~\citep{cheng2026reasoning,seedgrpo,sun2025ktaemodelfreealgorithmkeytokens,li2025attention,chen2025beyond,li2026outcome}. While effective, these methods often require auxiliary models, extra labels, or proxy signals that are only indirectly tied to environment feedback.

On-policy distillation provides another way to obtain dense supervision. OPD~\citep{opd} uses a stronger teacher to supervise the student's on-policy trajectories and provide token-level signals. However, it requires an additional stronger model, which increases computational cost and may introduce distribution mismatch. Related self-distillation methods condition the teacher on the student model and privileged signals, such as verifier feedback, future context, or correct trajectories~\citep{opsd,sdpo,sdft}, but they may still suffer from privileged information leakage. SDPO~\citep{sdpo} and RLSD~\citep{rlsd} further combine distillation signals with RL rewards to improve RL training while maintaining finer-grained credit assignment. However, these methods are mainly studied on simple reasoning tasks, leaving complex multi-turn, long-horizon agentic tasks underexplored.
\section{Method}
\begin{figure*}[h]
    \centering
    \includegraphics[width=\linewidth]
    {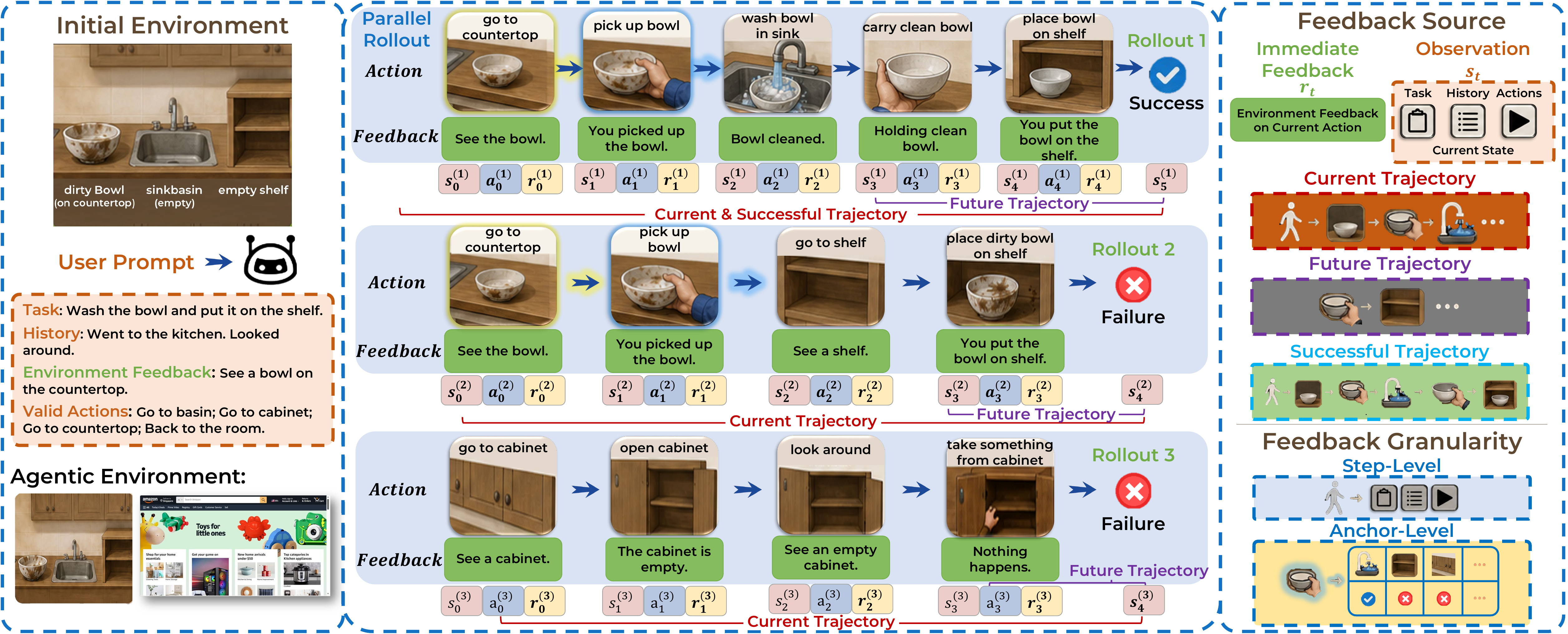}
    \caption{Pipeline of environment-feedback guided agent RL. The upper part summarizes training-only hindsight sources and placement choices. The lower part shows \method: placed feedback is exposed only to the teacher, which converts hindsight into reward-aligned action-level credit for GRPO.}
    \label{fig:pipeline}
    % \vspace{-1em}
\end{figure*}

\subsection{Preliminaries}
\paragraph{Problem setting.}
We consider a multi-turn LLM agent trajectory
\begin{equation}
    \tau=(s_0,a_0,r_0,s_1,a_1,r_1,\ldots,s_T,a_T,r_T),
\end{equation}
where $s_t$ is the environment state or observation before acting, $a_t$ is the executable action, and $r_t$ is the environment feedback returned after executing $a_t$. Each action is generated as a token sequence $a_t=(y_{t,1},\ldots,y_{t,L_t})$. Let $h_t=(s_0,a_0,r_0,\ldots,s_t)$ denote the history available when the agent produces $a_t$.

\paragraph{GRPO.}
GRPO~\citep{grpo} optimizes an on-policy group of rollouts without training an additional value model. For a task instance, let $\{\tau^n\}_{n=1}^{N}$ be trajectories sampled from the old policy $\pi_{\theta_{\mathrm{old}}}$ with outcome rewards $R^n$. GRPO computes a group-relative advantage
\begin{equation}
    A^n=\frac{R^n-\operatorname{mean}_{m}(R^m)}
    {\operatorname{std}_{m}(R^m)+\epsilon_A}.
\end{equation}
In long-horizon agent training, this reward-derived signal is usually broadcast to many tokens in the trajectory. We write $A_t$ for the advantage assigned to the action step $t$; for standard trajectory-level GRPO, $A_t=A^n$ for all steps in trajectory $\tau^n$. For token $y_{t,i}$, the policy ratio is
\begin{equation}
    \rho_{t,i}(\theta)=
    \frac{\pi_\theta(y_{t,i}\mid h_t,y_{t,<i})}
    {\pi_{\theta_{\mathrm{old}}}(y_{t,i}\mid h_t,y_{t,<i})}.
\end{equation}
For any token-level advantage $B_{t,i}$, the clipped GRPO surrogate is
\begin{equation}
    \ell_{\mathrm{GRPO}}(\theta;B_{t,i})
    =
    \min\left(
    \rho_{t,i}(\theta)B_{t,i},
    \operatorname{clip}(\rho_{t,i}(\theta),1-\epsilon,1+\epsilon)B_{t,i}
    \right).
\end{equation}
Standard GRPO optimizes this surrogate with $B_{t,i}=A_t$. This objective gives a stable reward direction, but its credit assignment is coarse: the same advantage can update decisive actions, reasoning tokens, formatting tokens, and repeated environment interactions.

\paragraph{On-policy distillation.}
On-policy distillation (OPD)~\citep{opd} provides the complementary signal. A teacher policy $\pi_T$ scores the student's sampled trajectory and supplies token-level supervision:
\begin{equation}
    \mathcal{L}_{\mathrm{OPD}}
    =
    \sum_{t,i}
    \operatorname{KL}
    \left[
    \pi_T(\cdot\mid h_t,y_{t,<i},z_t)
    \,\middle\|\,
    \pi_\theta(\cdot\mid h_t,y_{t,<i})
    \right],
\end{equation}
where $z_t$ denotes the additional context given to the teacher. In agent settings, useful $z_t$ often comes from environment hindsight, such as post-action feedback, future observations, or successful trajectories. This makes OPD dense, but also risky: if the teacher sees information unavailable to the student at decision time, directly imitating the teacher can leak privileged information, amplify stylistic teacher preferences, and destabilize training.

% \paragraph{Design principle.}
% \method combines these two signals asymmetrically. GRPO supplies the reliable update direction through $A_t$, while environment-conditioned distillation only adjusts where and how strongly this update is applied. In other words, \method does not treat the teacher as a full generative target. It uses the teacher to identify action tokens whose probabilities change under environment hindsight, then converts this evidence into a bounded reweighting of the GRPO advantage.

\subsection{Hindsight Placement for Agent Credit Assignment}

Environment hindsight is useful only when it is attached to the decisions it can explain. Since agent rollouts mix meaningful state changes with repeated or state-preserving interactions, \method separates the \emph{source} of feedback from its \emph{placement}, as summarized in Figure~\ref{fig:pipeline}.

Let $\mathcal{F}_t$ denote a training-only hindsight signal derived from the environment. Its source can be immediate feedback $r_t$, next observation $s_{t+1}$, future trajectory $\tau_{>t}$, successful trajectory $\tau^+$, current trajectory $\tau_{\leq t}$, or their combinations; Appendix~\ref{app:env_feedback} gives the detailed taxonomy. The student never receives this hindsight as decision-time input. A placement operator selects what the teacher sees:
\begin{equation}
    \Phi(t) = \operatorname{Place}(\mathcal{F}, t),
\end{equation}
where $\Phi(t)$ is the feedback used when scoring the sampled action at step $t$. This formulation lets the same source support different credit-assignment granularities.

\paragraph{Step-Level.}
The densest option is to attach feedback to every transition:
\begin{equation}
    \Phi_{\mathrm{step}}(t)=\mathcal{F}_t .
\end{equation}
The teacher scores each action token conditioned on local hindsight, $\pi_T(\cdot\mid h_t,y_{t,<i},\Phi_{\mathrm{step}}(t))$. This gives the densest credit signal and is useful when each transition contains locally causal information. However, repeated observations, failed interface attempts, or state-preserving moves can make dense placement noisy.

\paragraph{Anchor-Level.}
To reduce redundancy, we group decision steps into semantic anchors $\{\mathcal{C}_1,\ldots,\mathcal{C}_M\}$, where each anchor contains steps with the same or highly similar environment state. Let $g(t)=m$ if step $t$ belongs to anchor $\mathcal{C}_m$. The feedback attached to that anchor is
\begin{equation}
    \mathcal{F}^{A}_{m}
    =
    \operatorname{Agg}\left(\{\mathcal{F}_t:t\in\mathcal{C}_m\}\right),
    \qquad
    \Phi_{\mathrm{anchor}}(t)=\mathcal{F}^{A}_{g(t)} .
\end{equation}
Here $\operatorname{Agg}(\cdot)$ selects or merges feedback within the same semantic state. Anchor-level placement trades some density for robustness by concentrating hindsight on meaningful state changes. In the following, $\Phi(t)$ denotes either placement choice.

\subsection{Environment-Guided Advantage Reweighting}
\vspace{-1em}
\begin{figure*}[ht]
    \centering
    \includegraphics[width=\linewidth]
    {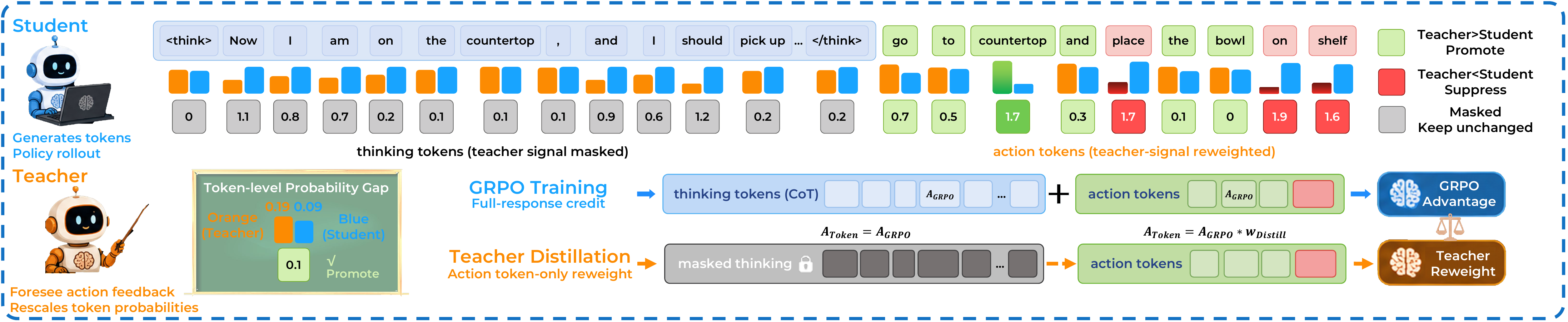}
    \caption{Mechanism of \method. A stop-gradient teacher conditioned on placed hindsight $\Phi(t)$ produces a teacher--student likelihood gap. \method signs this gap with the GRPO advantage, then clips, decays, and masks it to action tokens, so hindsight changes update magnitude and locality while reward determines direction.}
    \label{fig:method}
    % \vspace{-1em}
\end{figure*}

% \vspace{-1em}
After placing hindsight, the second question is how to use it without turning privileged information into a direct imitation target. \method converts teacher evidence into a reward-aligned coefficient rather than a standalone learning direction, as illustrated in Figure~\ref{fig:method}.

Let $\pi_T$ be a stop-gradient hindsight teacher. The student scores token $y_{t,i}$ using only the decision-time context $(h_t,y_{t,<i})$, while the teacher additionally receives placed environment feedback $\Phi(t)$. We define the teacher--student log-probability gap as
\begin{equation}
    \Delta_{t,i}
    =
    \log \pi_T(y_{t,i}\mid h_t,y_{t,<i},\Phi(t))
    -
    \log \pi_\theta(y_{t,i}\mid h_t,y_{t,<i}).
\end{equation}
If $\Delta_{t,i}>0$, the sampled token becomes more plausible after the teacher observes hindsight. Since this signal may also reflect privileged context or teacher style, we use it only to adjust the magnitude of the reward-driven update:
\begin{equation}
    w_{t,i}
    =
    \operatorname{clip}
    \left(
    \exp\left(\operatorname{sgn}(A_t)\operatorname{stopgrad}(\Delta_{t,i})\right),
    w_{\min}, w_{\max}
    \right).
\end{equation}
The sign of the GRPO advantage decides how teacher evidence is interpreted. For positive-advantage actions, teacher-supported tokens receive larger updates. For negative-advantage actions, teacher-supported tokens are penalized less aggressively, while teacher-disfavored tokens receive stronger penalties. Clipping prevents noisy hindsight probabilities from dominating reward learning, and the stop-gradient keeps the teacher signal as a coefficient rather than a hidden auxiliary objective.

We further restrict this reweighting to executable action tokens. Let $m^{\mathrm{act}}_{t,i}\in\{0,1\}$ indicate whether token $y_{t,i}$ belongs to the action span. We set
\begin{equation}
    \bar{w}_{t,i}
    =
    m^{\mathrm{act}}_{t,i}w_{t,i}
    +
    (1-m^{\mathrm{act}}_{t,i}),
\end{equation}
so reasoning and formatting tokens keep the original GRPO weight. This focuses hindsight on executable decisions, where environment feedback is most causally tied to task success. The final token advantage is
\begin{equation}
    \widetilde{A}_{t,i}
    =
    A_t\left((1-\alpha_k)+\alpha_k\bar{w}_{t,i}\right),
\end{equation}
where $\alpha_k\in[0,1]$ controls the strength of hindsight reweighting at training step $k$. We decay $\alpha_k$ over training: early updates exploit dense hindsight when exploration is weak, while later updates return control to environment rewards to reduce privileged-teacher bias.

\subsection{Selective Hindsight Objective}
The reweighted RL objective plugs $\widetilde{A}_{t,i}$ into the GRPO surrogate defined above:
\begin{equation}
    \mathcal{L}_{\mathrm{rw}}
    =
    -\sum_{t,i}
    \ell_{\mathrm{GRPO}}(\theta;\widetilde{A}_{t,i}).
\end{equation}
We additionally keep a lightweight action-only distillation term:
\begin{equation}
    \mathcal{L}_{\mathrm{act}}
    =
    \sum_{t,i}
    m^{\mathrm{act}}_{t,i}
    \operatorname{KL}
    \left[
    \pi_T(\cdot\mid h_t,y_{t,<i},\Phi(t))
    \,\middle\|\,
    \pi_\theta(\cdot\mid h_t,y_{t,<i})
    \right].
\end{equation}
This term targets the decision boundary of the agent, such as \texttt{search[...]} and \texttt{click[...]} in WebShop or environment actions in ALFWorld. It complements reweighting while avoiding full-trajectory distillation over long reasoning traces.

The final objective is
\begin{equation}
    \mathcal{L}_{\mathrm{SERL}}
    =
    \mathcal{L}_{\mathrm{rw}}
    +
    \lambda\mathcal{L}_{\mathrm{act}} .
\end{equation}
Thus, \method differs from direct OPD and loss-level RL--distillation mixtures: privileged hindsight never determines the full-response update direction. It enters through a placement operator, modifies only bounded action-level credit, and decays as the policy becomes more competent.

\section{Experiments}
\label{sec:exp}
\subsection{Experimental Setup}
\label{sec:exp_setup}
\paragraph{Benchmarks.}
We train LLM agents on two challenging benchmarks: ALFWorld~\citep{alfworld} and WebShop~\citep{webshop}.
\textit{ALFWorld} is an embodied environment for evaluating multi-step decision-making. In each episode, the agent receives a text goal and completes it through multi-turn interaction with the environment. The benchmark contains 3,827 task instances across six common household task categories: Pick \& Place (Pick), Examine in Light (Look), Clean \& Place (Clean), Heat \& Place (Heat), Cool \& Place (Cool), and Pick Two \& Place (Pick2).
\textit{WebShop} is a web-based interactive environment for evaluating agents in realistic online shopping scenarios. To complete each task, the agent interacts with a simulated HTML shopping website to search, navigate, and purchase a suitable item. The benchmark contains over 1.1 million products and 12k user instructions, providing a large and diverse action space.

\paragraph{Baselines.}
For ALFWorld and WebShop, we compare our approach with several competitive baselines. 
(1) \textit{Prompting methods}: ReAct~\citep{react} and Reflexion~\citep{reflexion}, which use in-context prompting to guide multi-step behavior without parameter updates. 
(2) \textit{RL training methods}: PPO~\citep{schulman2017proximal}, a standard actor-critic algorithm that requires an additional value model; critic-free group-based methods RLOO~\citep{rloo} and GRPO~\citep{grpo}, which estimate advantages from trajectory groups; and GIGPO~\citep{gigpo} and HGPO~\citep{hgpo}, which further perform step-level or hierarchical advantage estimation within trajectory groups. 
(3) \textit{RL--distillation hybrid methods}: SDPO~\citep{sdpo}, which uses environment feedback for self-distillation; two variants that combine SDPO with GRPO; and RLSD~\citep{rlsd}, which integrates self-distillation with token-level weighting in the RL objective.
% Full training settings and hyperparameter details are provided in Appendix~\ref{appendix:train_detail}.
\paragraph{Training Details.}
We use Qwen2.5-7B-Instruct~\cite{qwen2.5} as the base model. For ALFWorld and WebShop, all RL-based methods, including our approach and the baselines, are trained with the same hyperparameter settings for fair comparison. For group-based RL methods, the rollout group size $N$ is set to 8. The actor is optimized with a learning rate of $1\times10^{-6}$. We use PPO-style mini-batches of 256 samples and micro-batches of 32 samples for ALFWorld, and mini-batches of 64 samples and micro-batches of 8 samples for WebShop.
For RL methods combined with on-policy self-distillation (\method, RLSD~\cite{rlsd}), we set the initial self-distillation mixing coefficient to $\lambda=0.5$, linearly decay it over 50 steps, and clip the token-level distillation weights with a threshold of 0.2. The teacher policy is synchronized every 10 training steps.

\subsection{Experimental Results}

\begin{table}[t]
\centering
\caption{Main comparison of \method with representative baselines on ALFWorld and WebShop. For ALFWorld, we report the success rate (\%) for each subtask and the average success rate across all subtasks. For WebShop, we report the average score and success rate (\%). All self-distillation methods use immediate environment feedback $r_t$. $^{\dag}$ indicates results from prior reports.}
\label{tab:main_compare}
\resizebox{\textwidth}{!}{
\begin{tabular}{llccccccc|cc}
\toprule
\multirow{2}{*}{Type} & \multirow{2}{*}{Method} & \multicolumn{7}{c|}{\textbf{ALFWorld}} & \multicolumn{2}{c}{\textbf{WebShop}} \\
 & & Pick & Look & Clean & Heat & Cool & Pick2 & All & Score & Succ.\\
\midrule
Prompting$^{\dag}$ & Qwen2.5-7B-Instruct & 33.4 & 21.6 & 19.3 & 6.9 & 2.8 & 3.2 & 14.8 & 26.4 & 7.8\\
Prompting$^{\dag}$ & ReAct~\cite{react} & 48.5 & 35.4 & 34.3 & 13.2 & 18.2 & 17.6 & 31.2 & 46.2 & 19.5\\
Prompting$^{\dag}$ & Reflexion~\cite{reflexion} & 62.0 & 41.6 & 44.9 & 30.9 & 36.3 & 23.8 & 42.7 & 58.1 & 28.8\\
\midrule
RL Training$^{\dag}$ & PPO (with critic)~\cite{schulman2017proximal} & 92.3 & 64.0 & \textbf{92.5} & 89.5 & 80.3 & 68.8 & 80.4 & 81.4 & 68.7 \\
RL Training$^{\dag}$ & RLOO~\cite{rloo} & 87.6 & 78.2 & 87.3 & 81.3 & 71.9 & 48.9 & 75.5 & 80.3 & 65.7 \\
RL Training & GRPO~\cite{grpo} & 90.3 & 83.3 & 84.2 & 70.0 & 69.2 & 55.0 & 75.3 & 73.1 & 64.1 \\
RL Training & GIGPO~\cite{gigpo} & 93.5 & 83.3 & 78.9 & 86.7 & 76.2 & \textbf{85.0} & 83.9 & 83.5 & 75.8 \\
RL Training & HGPO~\cite{hgpo} & 92.3 & \underline{91.7} & 77.8 & \underline{93.3} & \textbf{85.7} & 73.9 & \underline{85.8} & \underline{88.4} & \underline{77.8} \\
Self-Distillation & SDPO~\cite{sdpo} & 66.7 & 66.7 & 29.6 & 16.7 & 20.0 & 0.0 & 33.3 & 0.0 & 0.0 \\
Hybrid & GRPO+SDPO(Advantage)~\cite{sdpo} & \textbf{100.0} & \underline{91.7} & \underline{88.9} & 53.3 & \underline{81.0} & 21.7 & 72.8 & 84.8 & 75.4 \\
Hybrid & GRPO+SDPO(Loss)~\cite{sdpo} & \underline{97.4} & \textbf{100.0} & \underline{88.9} & \textbf{100.0} & 71.4 & 34.8 & 82.1 & \underline{88.4} & 73.0 \\
Hybrid & RLSD~\cite{rlsd} & \underline{97.4} & 75.0 & \underline{88.9} & \textbf{100.0} & 61.9 & 73.9 & 82.9 & 83.6 & 75.8 \\
\midrule
Hybrid & SERL (Ours) & 92.3 & \textbf{100.0} & \underline{88.9} & \textbf{100.0} & 76.2 & \underline{82.6} & \textbf{90.0} & \textbf{89.5} & \textbf{80.1} \\
\bottomrule
\end{tabular}
}
\vspace{-1em}
\end{table}
\paragraph{Overall comparison.}
Table~\ref{tab:main_compare} compares \method with prompting agents, pure RL training methods, and RL--distillation hybrids on ALFWorld and WebShop. All self-distillation and hybrid variants in this comparison use the same immediate environment feedback $r_t$, so the key difference is how each method converts that feedback into an optimization signal. \method obtains the best aggregate performance on both benchmarks, reaching $90.0\%$ average success on ALFWorld, $89.5$ WebShop score, and $80.1\%$ WebShop success. Relative to GRPO, \method improves ALFWorld average success by $14.7$ points and WebShop success by $16.0$ points. Relative to the strongest pure RL baseline, HGPO, it still improves ALFWorld by $4.2$ points and WebShop success by $2.3$ points.

These gains indicate that the main bottleneck is not only variance reduction in policy-gradient estimation. HGPO and GIGPO already refine trajectory-level advantages, but their updates are still derived from sparse outcome rewards. \method instead uses environment feedback to reshape credit at the action-token level while keeping the reward-determined direction. This distinction matters in long-horizon agents: a successful or failed rollout contains many reasoning tokens, interface actions, and repeated observations, but only a small subset of executable actions changes the environment in a task-relevant way.

\method also outperforms OPD-style hybrids such as GRPO+SDPO and RLSD. These methods improve some individual ALFWorld categories, but their gains are less consistent across tasks. For example, GRPO+SDPO performs well on several short-horizon categories but drops sharply on Pick2, where the agent must preserve state across multiple object interactions. This pattern suggests that dense teacher supervision alone can over-emphasize local signals that are not causally aligned with final success. In contrast, \method uses the teacher asymmetrically: hindsight changes the magnitude and locality of the GRPO update, but the reward still determines whether the sampled behavior should be reinforced or suppressed. The aggregate improvement across ALFWorld and WebShop supports this controlled use of environment feedback.

\paragraph{Feedback sources.}

\begin{table}[t]
\centering
\caption{Analysis of \method with different feedback sources. We compare immediate environment feedback, next observations, future trajectories, successful trajectories, and their combinations.}
\label{tab:main_analyse}
\resizebox{\textwidth}{!}{
\begin{tabular}{lccccccc|cc}
\toprule
\multirow{2}{*}{Type} 
& \multicolumn{7}{c|}{\textbf{ALFWorld}} 
& \multicolumn{2}{c}{\textbf{WebShop}} \\
\cmidrule(lr){2-8} \cmidrule(lr){9-10}
& Pick & Look & Clean & Heat & Cool & Pick2 & All & Score & Succ. \\
\midrule
GRPO~\cite{grpo} 
& 90.3 & 83.3 & 84.2 & 70.0 & 69.2 & 55.0 & 75.3 & 73.1 & 64.1 \\
GIGPO~\cite{gigpo} 
& 93.5 & 83.3 & 78.9 & 86.7 & 76.2 & \textbf{85.0} & 83.9 & 83.5 & 75.8 \\
HGPO~\cite{hgpo} 
& 92.3 & \underline{91.7} & 77.8 & \underline{93.3} & \underline{85.7} & 73.9 & 85.8 & 88.4 & 77.8 \\
\midrule
immediate feedback
& 92.3 & \textbf{100.0} & 88.9 & \textbf{100.0} & 76.2 & \underline{82.6} & \underline{90.0} & \underline{89.5} & \underline{80.1} \\
next observation 
& \underline{97.4} & 72.2 & \underline{91.7} & 80.0 & 61.9 & 56.5 & 76.6 & \textbf{90.5} & 77.7 \\
future trajectory 
& \textbf{100.0} & 83.3 & 83.3 & 86.7 & 80.1 & 65.2 & 83.1 & 85.9 & 76.6 \\
successful trajectory or immediate feedback
& 97.0 & \textbf{100.0} & 83.3 & 73.3 & \underline{85.7} & 52.2 & 81.9 & 84.1 & 72.7 \\
successful trajectory and immediate feedback
& \textbf{100.0} & \textbf{100.0} & \textbf{94.4} & \underline{93.3} & 81.0 & 73.9 & \textbf{90.4} & 87.7 & \textbf{81.3} \\
successful trajectory and next observation 
& 94.9 & \underline{91.7} & 88.9 & \textbf{100.0} & \underline{85.7} & 52.2 & 85.6 & 86.9 & 77.7 \\
successful trajectory and future trajectory 
& \textbf{100.0} & 83.3 & 83.3 & 73.3 & \textbf{90.5} & 69.6 & 83.3 & 88.1 & 76.6 \\
successful trajectory, future trajectory, and immediate feedback
& 90.0 & 91.6 & 83.3 & 66.7 & \underline{85.7} & 39.1 & 76.1 & 87.1 & 77.0 \\
successful trajectory, future trajectory, and next observation 
& \textbf{100.0} & 83.3 & 83.3 & 60.0 & 81.0 & 52.2 & 76.6 & 84.1 & 76.2 \\
\bottomrule
\end{tabular}
}
\vspace{-1em}
\end{table}
Table~\ref{tab:main_analyse} studies the feedback-source axis defined in our method. The central observation is that more privileged context is not monotonically better. Immediate feedback $r_t$ is already a strong source, achieving $90.0\%$ ALFWorld average success and $80.1\%$ WebShop success. Its strength comes from locality: it is produced immediately after the current action, so the teacher can connect probability changes to the environment's response without relying on a long future trajectory.

The best overall source combines successful trajectories with immediate feedback, reaching the highest ALFWorld average ($90.4\%$) and the highest WebShop success ($81.3\%$). This combination is informative because the two signals play different roles. A successful trajectory supplies a positive behavioral reference, while $r_t$ anchors that reference to the current rollout's actual transition. Without this grounding, successful trajectories alone or disjunctive combinations with immediate feedback are weaker, suggesting that reference behavior must be tied to the state being credited rather than used as a generic demonstration.

The weaker results for next observations, future trajectories, and large source combinations further clarify what makes feedback useful. Next observations achieve the highest WebShop score ($90.5$) but lower success, indicating that rich post-action state information can improve partial progress or item ranking without reliably improving final completion. Future trajectories and multi-source privileged feedback often underperform because they contain delayed consequences that are not uniquely attributable to the current action. Thus, effective agent feedback should be local enough to preserve causal alignment and rich enough to reveal why an action changed the environment; simply giving the teacher more hindsight can introduce noise and privileged mismatch.

\paragraph{Feedback placement.}
\begin{table}[t]
\centering
\caption{Analysis of feedback granularity in \method on WebShop. Step-level feedback applies distillation to every transition, whereas anchor-level feedback first groups semantically similar environment states before applying distillation.}
\label{tab:main_anchor}
\resizebox{\textwidth}{!}{
\begin{tabular}{lcc|cc}
\toprule
\multirow{2}{*}{Type} 
& \multicolumn{2}{c|}{\textbf{Step Level}} 
& \multicolumn{2}{c}{\textbf{Anchor Level}} \\
\cmidrule(lr){2-3} \cmidrule(lr){4-5}
& Score & Succ. & Score & Succ. \\
\midrule
immediate feedback & 89.5 & 80.1 & \textbf{91.5} & 79.3 \\
next observation & \textbf{90.5} & 77.7 & 87.2 & 76.2 \\
future trajectory & 85.9 & 76.6 & 83.2 & 75.4 \\
successful trajectory or immediate feedback & 84.1 & 72.7 & 87.1 & 74.8 \\
successful trajectory and immediate feedback & 87.7 & \textbf{81.3} & 88.1 & \textbf{81.9} \\
successful trajectory and next observation & 86.9 & 77.7 & 81.8 & 79.7 \\
successful trajectory and future trajectory & 88.1 & 76.6 & 90.5 & 77.7 \\
successful trajectory, future trajectory, and immediate feedback & 87.1 & 77.0 & 89.2 & 79.3 \\
successful trajectory, future trajectory, and next observation & 82.0 & 72.9 & 85.5 & 76.3 \\
\bottomrule
\end{tabular}
}
\vspace{-1.5em}
\end{table}
Table~\ref{tab:main_anchor} evaluates the placement axis on WebShop. Step-level placement maximizes density by applying feedback to every transition, whereas anchor-level placement groups semantically similar states and applies feedback around meaningful state changes. The results show that placement should be chosen together with the feedback source. With immediate feedback, anchor-level placement improves score from $89.5$ to $91.5$, suggesting that grouping suppresses redundant updates while preserving the useful local signal. With successful trajectory plus immediate feedback, anchor-level placement further improves success from $81.3\%$ to $81.9\%$, the best WebShop success rate in the table.

Anchor-level placement is not uniformly superior, and this is also informative. For next observations and future trajectories used alone, step-level placement gives higher score and success, likely because these signals are path-specific: grouping similar-looking states can remove fine-grained temporal information that explains the current action. By contrast, anchor-level placement is more helpful for multi-source or partially privileged feedback, where raw density can amplify redundant or weakly causal teacher evidence. This supports the main design principle of \method: dense environment feedback is valuable only when its source and insertion point are aligned with the decisions that actually affect the future trajectory.

\subsection{Ablation Study}
\paragraph{LLM-judged feedback.}
\begin{table}[t]
\centering
\caption{Performance on \method using LLM-judged feedback. We compare trajectory judgments generated by Kimi-K2.6 and Qwen2.5-7B-Instruct, using either the current trajectory alone or together with a successful trajectory.}
\label{tab:llm_judge}
\resizebox{\textwidth}{!}{
\begin{tabular}{llccccccc|cc}
\toprule
\multirow{2}{*}{Model} 
& \multirow{2}{*}{Feedback Source} 
& \multicolumn{7}{c|}{\textbf{ALFWorld}} 
& \multicolumn{2}{c}{\textbf{WebShop}} \\
\cmidrule(lr){3-9} \cmidrule(lr){10-11}
& & Pick & Look & Clean & Heat & Cool & Pick2 & All & Score & Succ. \\
\midrule
\multirow{2}{*}{Kimi-K2.6} 
& Current Trajectory 
& 97.4 & 75.0 & 88.9 & \textbf{100.0} & 76.2 & \textbf{82.6} & 86.7 & 88.2 & 78.1 \\
& Current Trajectory + Successful Trajectory 
& 90.0 & \textbf{100.0} & 88.9 & 93.3 & \textbf{90.5} & 73.9 & \textbf{89.4} & \textbf{89.0} & \textbf{81.8} \\
\midrule
\multirow{2}{*}{Qwen2.5-7B-Instruct} 
& Current Trajectory 
& 97.4 & 58.3 & \textbf{94.4} & 66.7 & 76.2 & 65.2 & 76.4 & 81.1 & 69.2 \\
& Current Trajectory + Successful Trajectory 
& \textbf{100.0} & 58.3 & 77.8 & 73.3 & 81.0 & 69.6 & 76.7 & 82.0 & 68.8 \\
\bottomrule
\end{tabular}
}
\vspace{-1em}
\end{table}

Table~\ref{tab:llm_judge} tests whether raw environment trajectories can be compressed into a higher-level feedback summary by an external LLM judge. This ablation probes a different form of feedback reuse: instead of giving the teacher raw observations or trajectories, a judge first converts the rollout into a concise diagnosis. With Kimi-K2.6, current-trajectory judgments improve substantially over GRPO, and adding a successful trajectory raises WebShop success to $81.8\%$. This shows that a capable judge can turn a long observation--action sequence into action-relevant credit: it identifies the causal error or useful behavior and exposes it to the teacher as compact privileged guidance.

The judge model itself becomes part of the feedback channel. Qwen2.5-7B-Instruct provides some useful signal, especially on ALFWorld, but is much weaker on WebShop. The likely reason is context coverage. ALFWorld observations are short, whereas WebShop trajectories contain long HTML-like observations and action histories. In our setting, the Qwen judge has a 32K context window, which can truncate or compress away important WebShop evidence, while Kimi-K2.6 supports a 256K context window and can usually cover the complete trajectory prompt. The result highlights a practical constraint: LLM-judged feedback is useful only when the judge has enough context length and reasoning capacity to preserve the causal history being summarized.

\paragraph{Decay of teacher signal.}
\begin{wrapfigure}{r}{0.52\linewidth}
    \vspace{-1.0em}
    \centering
    \includegraphics[width=\linewidth]{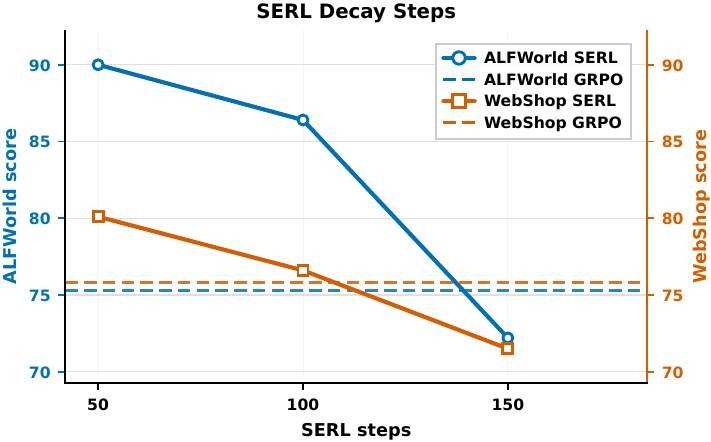}
    \caption{Effect of decaying the hindsight teacher signal during training. \textbf{Left (a)} shows reward score, \textbf{Middle (b)} shows policy entropy, and \textbf{Right (c)} shows response length. Decay uses dense teacher feedback early for credit assignment and gradually returns optimization control to reward-driven GRPO.}
    \label{fig:decay}
    % \vspace{-1em}
\end{wrapfigure}

Figure~\ref{fig:decay} studies the temporal role of the teacher-induced reweighting signal. The motivation follows the same principle as RLSD~\citep{rlsd}: teacher probabilities can provide dense magnitude information, but the direction of policy improvement should remain reward-driven. This concern is stronger in our setting because the teacher may condition on environment feedback, future context, or successful trajectories that the student will not observe at test time.

The training curves show why a decayed teacher weight is preferable to using hindsight uniformly throughout training. Early in training, the policy explores poorly and sparse rewards provide weak credit assignment; teacher-conditioned reweighting helps identify which action tokens deserve larger updates. Later, as the policy becomes competent, the teacher--student gap increasingly mixes useful credit with privileged-context bias, formatting differences, and trajectory-specific noise. Decay therefore implements a staged use of feedback: hindsight accelerates early credit assignment, then optimization gradually returns to GRPO so that the final policy is anchored by environment rewards rather than persistent privileged supervision.

% \vspace{-1em}
\section{Conclusion}
% \vspace{-1em}
We studied credit assignment in long-horizon LLM agents from the perspective of environment feedback. The core challenge is that RL sparse rewards provide a reliable optimization direction but weak credit assignment, whereas hindsight distillation provides dense token-level signals but may rely on information unavailable to the student during training, potentially causing privileged-information leakage and training instability. \method{} addresses this challenge by using reward to determine the update direction and using environment-conditioned teacher signals only for bounded, action-level credit adjustment.
The broader lesson is that environment feedback should not be treated as unrestricted auxiliary supervision. Effective agent RL requires aligning three design choices: what feedback source is used, where it is inserted, and how strongly it is allowed to affect the policy update. Our results show that grounded, action-relevant feedback can be more useful than richer but weakly causal hindsight, and that semantic feedback placement can matter as much as feedback density. We hope this work provides new insights for the community and lays a foundation for future research on integrating environment feedback into the training of agentic LLMs.

\clearpage
\bibliographystyle{plainnat}
\bibliography{refs}

\clearpage
\appendix
\newpage
\appendix
\section{Environment Feedback Details}
\label{app:env_feedback}
We formalize a multi-turn agent trajectory as an alternating sequence of states, actions, and environment feedback:
\begin{equation}
    \tau = (s_0, a_0, r_0, s_1, a_1, r_1, \ldots, s_T, a_T, r_T),
\end{equation}
where $s_t$ denotes the environment state or observation available before action $a_t$, and $r_t$ denotes the feedback returned by the environment after executing $a_t$. In our analysis, environment feedback has two independent design axes: the source of feedback information and the position at which this information is injected into training.

\subsection{Feedback Sources}
We consider five feedback sources, ordered from local to increasingly privileged information.
\paragraph{Environment feedback.}
The most direct signal is the immediate environment feedback $r_t$ returned after the current action. It captures whether the action changes the environment in a useful way, but does not expose future decisions.

\paragraph{Next observation.}
The next observation $s_{t+1}$ provides the post-action state induced by $a_t$. Compared with $r_t$, it contains richer state information and can reveal how the environment actually responds to the current action.

\paragraph{Future trajectory.}
Future trajectory feedback uses the suffix after the current step,
\begin{equation}
    \tau_{>t} = \{(s_j, a_j)\}_{j>t}.
\end{equation}
This source provides delayed consequences of the current decision and can offer stronger supervision, but it is also more privileged because it contains information unavailable when $a_t$ is chosen.

\paragraph{Successful trajectory.}
A successful trajectory $\tau^{+}=(s^{+}_0,a^{+}_0,\ldots,s^{+}_{T},a^{+}_{T})$ is a rollout that reaches the task goal. It serves as a positive behavioral reference and can indicate which states or actions are associated with successful completion.

\paragraph{Current trajectory.}
The current trajectory prefix
\begin{equation}
    \tau_{\leq t} = (s_0,a_0,\ldots,s_t,a_t)
\end{equation}
contains only information produced by the agent so far. It is less privileged than future or successful trajectories, and mainly provides context for judging whether the current action is consistent with the agent's own interaction history.

\subsection{Temporal Granularity of Feedback}
We further distinguish where feedback is applied during training.
\paragraph{Step-level feedback.}
At the step level, feedback is injected independently for every transition $(s_t,a_t)$. This provides dense supervision and assigns credit to each action, but may also update many trivial or repetitive steps.

\paragraph{Anchor-level feedback.}
At the anchor level, transitions in a rollout group are first partitioned according to their environment states. An anchor contains steps whose states $s_t$ correspond to the same environment condition or are semantically similar. Feedback is then applied at the anchor rather than individual-step level. This reduces noisy updates on redundant interactions while preserving supervision for meaningful state changes.

% \section{Training Details}
% \label{appendix:train_detail}
% We use Qwen2.5-7B-Instruct~\cite{qwen2.5} as the base model. For ALFWorld and WebShop, all RL-based methods, including our approach and the baselines, are trained under the same hyperparameter settings for fair comparison. For group-based RL methods, the rollout group size $N$ is set to 8. The actor is optimized with a learning rate of $1\times10^{-6}$, using PPO-style mini-batches of 256 samples and per-GPU micro-batches of 32 samples.
% For RL methods combined with on-policy self-distillation, we set the initial self-distillation mixing coefficient to $\lambda=0.5$, linearly decay it over 50 steps, and clip the token-level distillation weights with a threshold of 0.2. The teacher policy is synchronized every 10 training steps.

\section{Training Details}
\subsection{Training Setup Details}
\label{apd:training_details}
We use Qwen2.5-7B-Instruct~\citep{qwen2.5} as the base model and conduct all experiments with the VeRL codebase~\citep{verl} on 8 NVIDIA H200 GPUs. For ALFWorld and WebShop, all RL-based methods, including our approach and the baselines, are trained with the same hyperparameter settings for fair comparison~\cite{gigpo}. For group-based RL methods, the rollout group size $N$ is set to 8. The actor is optimized for one training epoch with a learning rate of $1\times10^{-6}$ with total 150 steps. We set the maximum number of agent--environment interaction turns to 50 for ALFWorld and 15 for WebShop. For ALFWorld, we use PPO-style mini-batches of 256 samples and micro-batches of 32 samples; for WebShop, we use mini-batches of 64 samples and micro-batches of 8 samples. The rollout temperature is set to 0.4.

For RL methods combined with on-policy self-distillation, including \method{}, SDPO~\cite{sdpo} and RLSD~\citep{rlsd}, we set the initial self-distillation mixing coefficient to $\lambda=0.5$, linearly decay it over 50 steps, and clip the token-level distillation weights with a threshold of 0.2. The teacher policy is synchronized every 10 training steps.

\subsection{Reward Score}
\begin{figure*}[ht]
    \centering
    \includegraphics[width=\linewidth]
    {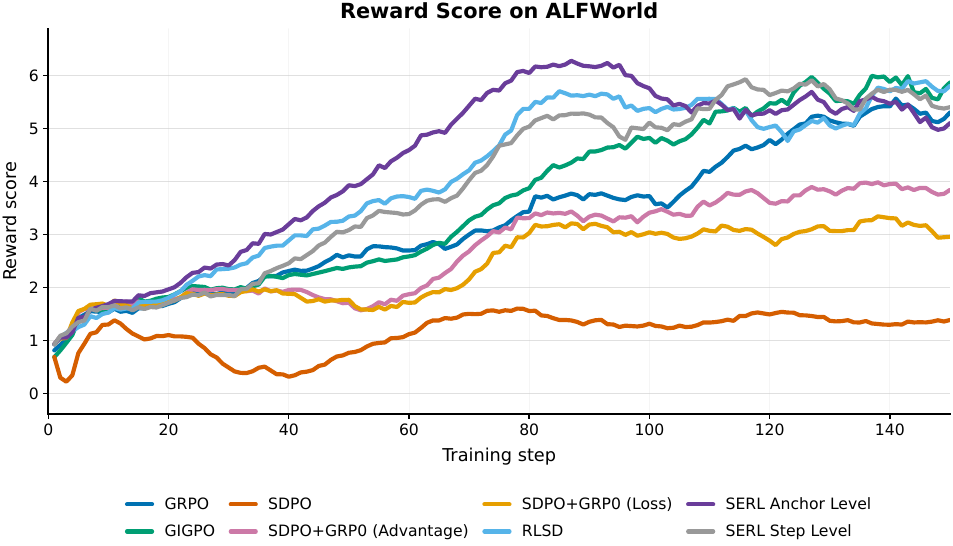}
    \caption{Training reward dynamics on agent environments. \method, which combines GRPO with environment-feedback-guided OPD, improves reward faster than pure RL baselines. Anchor-level feedback further accelerates convergence over step-level feedback by concentrating updates on semantically meaningful state changes.}
    \label{fig:rewardplot}
    % \vspace{-2em}
\end{figure*}

Figure~\ref{fig:rewardplot} provides the training reward dynamics behind the main results. The central observation is that combining GRPO with OPD-style environment feedback converges faster than pure RL training. This supports the motivation of \method: environment feedback is not merely an additional supervision source, but a dense credit signal that helps the policy identify which tokens and actions should receive stronger updates before sparse trajectory rewards become reliable.

The comparison between step-level and anchor-level feedback further clarifies where this dense signal should be applied. Step-level feedback gives every transition a teacher-conditioned update, which increases supervision density but also spends capacity on repeated observations, formatting tokens, and low-level interface operations. Anchor-level feedback instead groups semantically similar environment states and applies the signal at meaningful state changes. Its faster reward growth suggests that the main bottleneck is not the amount of feedback, but the precision of credit placement: agent training benefits most when environment feedback is concentrated on decision points that actually change the future trajectory.

\subsection{LLM Judge Prompt}

LLM-judged feedback is used to test a more general form of environment feedback summary. Instead of directly exposing raw observations or full trajectories to the teacher, a judge model reads the rollout and produces a short diagnostic summary. This matters for long-horizon agents because raw trajectories can be long, noisy, and hard to align token by token. A good judge converts the trajectory into an actionable credit signal: it states whether the rollout succeeded, identifies the causal error or useful behavior, and provides concise privileged guidance for the teacher. The ablation in Table~\ref{tab:llm_judge} therefore measures not only whether LLM judgments are helpful, but also whether the judge has enough context capacity to faithfully summarize the environment interaction.

\begin{center}
\fcolorbox{black!25}{black!3}{%
\begin{minipage}{0.94\linewidth}
\small
\textbf{Trajectory Judge System Prompt}

\vspace{0.35em}
\textit{Role.} You are an expert rollout critic.

\vspace{0.35em}
\textit{Input.} Given a full trajectory and optional reference trajectories, write a concise trajectory judgment that another policy model can use as privileged guidance.

\vspace{0.35em}
\textit{Requirements.}
The first sentence must explicitly state whether the rollout succeeded or failed. Summarize what went right or wrong, connect it to the task objective, and end with the most actionable lesson. Keep the judgment short and concrete. Always write at least one concrete sentence. Never answer with ``None'', ``N/A'', or only the section title.
\end{minipage}}
\end{center}

\subsection{Computational Cost}
\begin{wraptable}{r}{0.48\textwidth}
\centering
\caption{Computational cost of different training algorithms. MFU denotes model FLOPs utilization, and time per step is measured under the same training infrastructure.}
\label{tab:training_efficiency}
\resizebox{0.48\textwidth}{!}{
\begin{tabular}{lcc}
\toprule
Method & MFU & Time per Step (s) \\
\midrule
GRPO & 0.3625 & 255.5 \\
GIGPO & 0.3756 & 295.8 \\
SDPO+GRPO & 0.2877 & 282.3 \\
RLSD & 0.2928 & 255.6 \\
SERL (Step Level) & 0.2907 & 272.5 \\
SERL (Anchor Level) & 0.2462 & 295.9 \\
\bottomrule
\end{tabular}
}
\end{wraptable}

Table~\ref{tab:training_efficiency} compares the training cost of representative RL and RL--distillation algorithms under the same infrastructure. Pure RL methods such as GRPO and GIGPO achieve higher MFU because their updates are dominated by standard policy forward--backward computation. In contrast, distillation-based methods introduce additional teacher scoring, feedback construction, and token-level weighting, which lower MFU by adding non-matrix-multiplication overhead and more irregular sequence processing. This is expected in agent training, where environment interaction and trajectory-dependent feedback make the workload less uniform than standard supervised or reasoning-only RL.

The key observation is that \method remains in the same wall-clock cost regime as strong baselines while providing denser credit assignment. Step-level \method increases time per step only moderately over GRPO and is faster than SDPO+GRPO and GIGPO, suggesting that action-only reweighting keeps the additional teacher signal relatively lightweight. Anchor-level \method has higher per-step overhead because it performs semantic grouping and feedback aggregation before applying the teacher signal, but this cost buys a different trade-off: it spends computation on more meaningful state changes rather than uniformly supervising every transition. Thus, the relevant efficiency question for long-horizon agents is not only per-step throughput, but whether each update places credit on decisions that actually affect future interaction.

\section{Limitations}
\label{apd:limitation}
Our experiments are limited to two agentic environments, ALFWorld and WebShop, which may constrain the generalizability of the empirical findings. Extending on-policy training to more diverse agentic environments remains challenging, as it requires substantial computational resources and robust infrastructure for stable training and evaluation. We leave the broader-scale evaluation of \method across additional agentic environments as an important direction for future work.

% \section{Broader Impact}
% \label{apd:broader_impact}
% Our work aims to improve the training of long-horizon LLM agents by using environment feedback for more precise credit assignment. By enabling more efficient and effective learning, our method may support the development of intelligent agents for complex tasks in domains such as customer service, personal assistance, and workflow automation. At the same time, stronger agent training methods may also amplify potential risks. More capable agents could be misused to generate harmful content, automate unethical behavior, or execute unsafe actions at scale. These risks highlight the need for careful deployment, robust safety constraints, human oversight, and misuse-prevention mechanisms to ensure that such systems are developed and used responsibly.

% \clearpage
% \input{checklist.tex}

\end{document}